\PassOptionsToPackage{unicode}{hyperref}
\PassOptionsToPackage{hyphens}{url}
\documentclass[
  10pt,
]{article}
\usepackage{xcolor}
\usepackage{amsmath,amssymb}
\usepackage{iftex}
\ifPDFTeX
  \usepackage[T1]{fontenc}
  \usepackage[utf8]{inputenc}
  \usepackage{textcomp} %
\else %
  \usepackage{unicode-math} %
  \defaultfontfeatures{Scale=MatchLowercase}
  \defaultfontfeatures[\rmfamily]{Ligatures=TeX,Scale=1}
\fi
\usepackage{lmodern}
\ifPDFTeX\else
\fi
\IfFileExists{upquote.sty}{\usepackage{upquote}}{}
\IfFileExists{microtype.sty}{%
  \usepackage[]{microtype}
  \UseMicrotypeSet[protrusion]{basicmath} %
}{}
\makeatletter
\@ifundefined{KOMAClassName}{%
  \IfFileExists{parskip.sty}{%
    \usepackage{parskip}
  }{%
    \setlength{\parindent}{0pt}
    \setlength{\parskip}{6pt plus 2pt minus 1pt}}
}{%
  \KOMAoptions{parskip=half}}
\makeatother
\usepackage{longtable,booktabs,array}
\usepackage{calc} %
\usepackage{etoolbox}
\makeatletter
\patchcmd\longtable{\par}{\if@noskipsec\mbox{}\fi\par}{}{}
\makeatother
\IfFileExists{footnotehyper.sty}{\usepackage{footnotehyper}}{\usepackage{footnote}}
\makesavenoteenv{longtable}
\providecommand{\tightlist}{%
  \setlength{\itemsep}{0pt}\setlength{\parskip}{0pt}}
\usepackage{iclr2027_conference,times}
\iclrfinalcopy
\usepackage{graphicx}
\usepackage{float}
\usepackage{caption}
\usepackage{bookmark}
\IfFileExists{xurl.sty}{\usepackage{xurl}}{} %
\hypersetup{
  pdftitle={The Decomposition Tax: LLM Pipelines Lose Up to 40 Accuracy Points at Their Own Interfaces},
  hidelinks,
  pdfcreator={LaTeX via pandoc}}

\title{The Decomposition Tax: LLM Pipelines Lose Up to 40 Accuracy
Points at Their Own Interfaces}
\author{Tianqi Bu$^{1,2}$\thanks{Corresponding author: \texttt{tianqi.bu@rutgers.edu}} \quad
YuXuan Peng$^{3}$ \quad Junteng Tu$^{4}$ \quad Henghui Xiao$^{1}$ \\
$^{1}$Rutgers University \quad $^{2}$Nanjing Tech University \\
$^{3}$Nanjing University of Posts and Telecommunications \quad $^{4}$New York University}

\date{}

\begin{document}
\maketitle
\begin{abstract}
A four-stage LLM pipeline gives up as much as 40.5 accuracy points at
its own interfaces (gemma-3-12B on MATH-500, Holm-corrected p =
1.66e-19; the largest tax in the primary family). We hold model,
problem, stages, stage prompts and completion budget fixed, vary only
whether each stage can still see the original problem, and call the
accuracy difference the decomposition tax. Across 21 open-weight models
from nine organisations, on GSM-Hard and MATH-500 at n = 200 paired
items per cell, 70 of 118 primary-family tests survive
Benjamini-Hochberg correction and 54 survive Holm. On GSM-Hard, a
placebo recovers nothing: it carries at least 60\% of the extra tokens
and at most one word of the problem. Builders design a pipeline one
stage at a time, and its bill arrives at the interfaces between stages.
Rewriting one stage's instruction moves gemma-3-12B's tax from 4.5 to
36.5 points, and adding ``every relationship stated between them'' to a
stage that lists the numerical quantities lowers the tax on 9 of 9
models on MATH-500. Re-grounding, which shows a stage the original
problem again, belongs after the loss. With one lossy interface,
re-grounding the stage after it beats re-grounding the stage before it
on 7 of 7 models on both benchmarks; on MATH-500 the earlier repair is
worse than none on 7 of 7. Newer models still pay: gemma-4-12B gives up
37.0 points, and the repair holds on all three of the newest models we
test. A sealed held-out test refuted a stronger rule we registered,
which predicted the paying stage from the interface and receiver types,
so we locate the tax by measuring one stage at a time. The prescription
has two parts: re-ground the stage after the lossy interface, and if a
stage must list the quantities, tell it to keep the relationships.
\end{abstract}

\section{1. Introduction}\label{introduction}

Multi-stage LLM pipelines build one answer from several calls, and they
differ in how much each call still sees. Least-to-most prompting
\citep{zhou2022leastt} carries every earlier answer forward, while
Decomposed Prompting \citep{khot2022decomp} hands each sub-task handler
only its sub-query. A pipeline enters the one-message regime whenever a
context budget or a role boundary lets only a single message cross an
interface: each later stage then reasons only from what an earlier stage
chose to write down. This paper prices that regime.

Builders design a pipeline one stage at a time, and its bill arrives at
the interfaces between stages. The frameworks we know of let a builder
rewrite a stage's prompt, and none puts a price on the interface between
two stages. Failure studies catalogue and repair what goes wrong at that
interface \citep{cemri2025why,lin2026agentask}. The closest measurement
of its cost \citep{radhakrishnan2023questio} changes both the call
structure and what each stage can see. We price the interface alone.

We call that price the \textbf{decomposition tax}. In the
\texttt{strict} arm each stage sees only the previous stage's output. In
the \texttt{full} arm we re-ground each stage: it also sees the original
problem. \texttt{TAX\ =\ acc(full)\ $-$\ acc(strict)}, paired per item, is
what re-grounding buys back, and so what the interfaces were costing.
Both arms share model, stages, stage prompts and completion cap. A
single chain-of-thought call on the same model, the monolith, is the
reference. On MATH-500, gemma-3-12B solves 15.0\% of problems in the
strict arm and 55.5\% in the full arm. Strong models pay too:
gemma-4-12B, which solves 86.5\% of MATH-500 in one call, gives up 37.0
points. Re-grounding one stage alone, arm \texttt{g\_i}, measures how
much of the tax falls on stage \emph{i}, its \textbf{incidence}:
\texttt{STAGE\_TAX(i)\ =\ acc(g\_i)\ $-$\ acc(strict)}.

Two 2$\times$2 squares take one interface apart. A three-stage square crosses
what the sender keeps (CONTENT) with how it writes (FORM); a four-stage
square crosses the sender's whole instruction (EDGE) with what the
receiver may do (RECEIVER). Two mixed-order pipelines place one lossy,
condensing interface early or late, which separates the interface from
stage position. Figure 1 summarises the instrument and the findings.

\textbf{Contributions.}

\begin{enumerate}
\def\labelenumi{\arabic{enumi}.}
\tightlist
\item
  \textbf{A price tag on an interface.} The tax reaches \textbf{40.5
  accuracy points}. On the four-stage headline pipeline its bootstrap
  interval excludes zero for \textbf{17 of 20} models on MATH-500 and
  \textbf{16 of 20} on GSM-Hard, and \textbf{70 of 118} primary-family
  tests survive Benjamini-Hochberg correction (\textbf{54} survive
  Holm). Where a single call beats the strict pipeline by more than
  0.05, re-grounding recovers a median \textbf{63\%} of the gap (Section
  3.1).
\item
  \textbf{The tax is the missing problem.} On GSM-Hard, a placebo
  recovers nothing: it carries at least 60\% of re-grounding's extra
  tokens and at most one word of the problem. Sampling, parsing and
  hardware leave the tax in place, as does removing mathematical
  vocabulary from a four-stage pipeline (Section 3.1 and Section 3.2).
\item
  \textbf{One stage's instruction sets the tax.} Rewriting one stage's
  instruction moves gemma-3-12B's tax from 4.5 to 36.5 points, and
  asking a list-writing stage to keep ``every relationship stated
  between them'' lowers the tax on 9 of 9 models on MATH-500 (Section
  3.2).
\item
  \textbf{With one lossy interface, the repair belongs after it.}
  Re-grounding the stage after the lossy interface beats re-grounding
  the stage before it on 7 of 7 models on both benchmarks; on MATH-500
  the earlier repair is worse than none on 7 of 7 (Section 3.3).
\item
  \textbf{Newer models still pay, and the repair still works.} The three
  newest models we test all pay, gemma-4-12B the most at 37.0 points,
  and the repair holds on all three while, on MATH-500, the tax's split
  between edge and receiver shifts (Section 3.4).
\end{enumerate}

We score dated, registered predictions and report their verdicts
(Appendix~D, Appendix~F and Appendix~G).

\section{2. Setup}\label{setup}

\subsection{2.1 The arms differ only in what each stage
sees}\label{the-arms-differ-only-in-what-each-stage-sees}

\begin{figure}[t]
\centering
\includegraphics[width=1.00\linewidth]{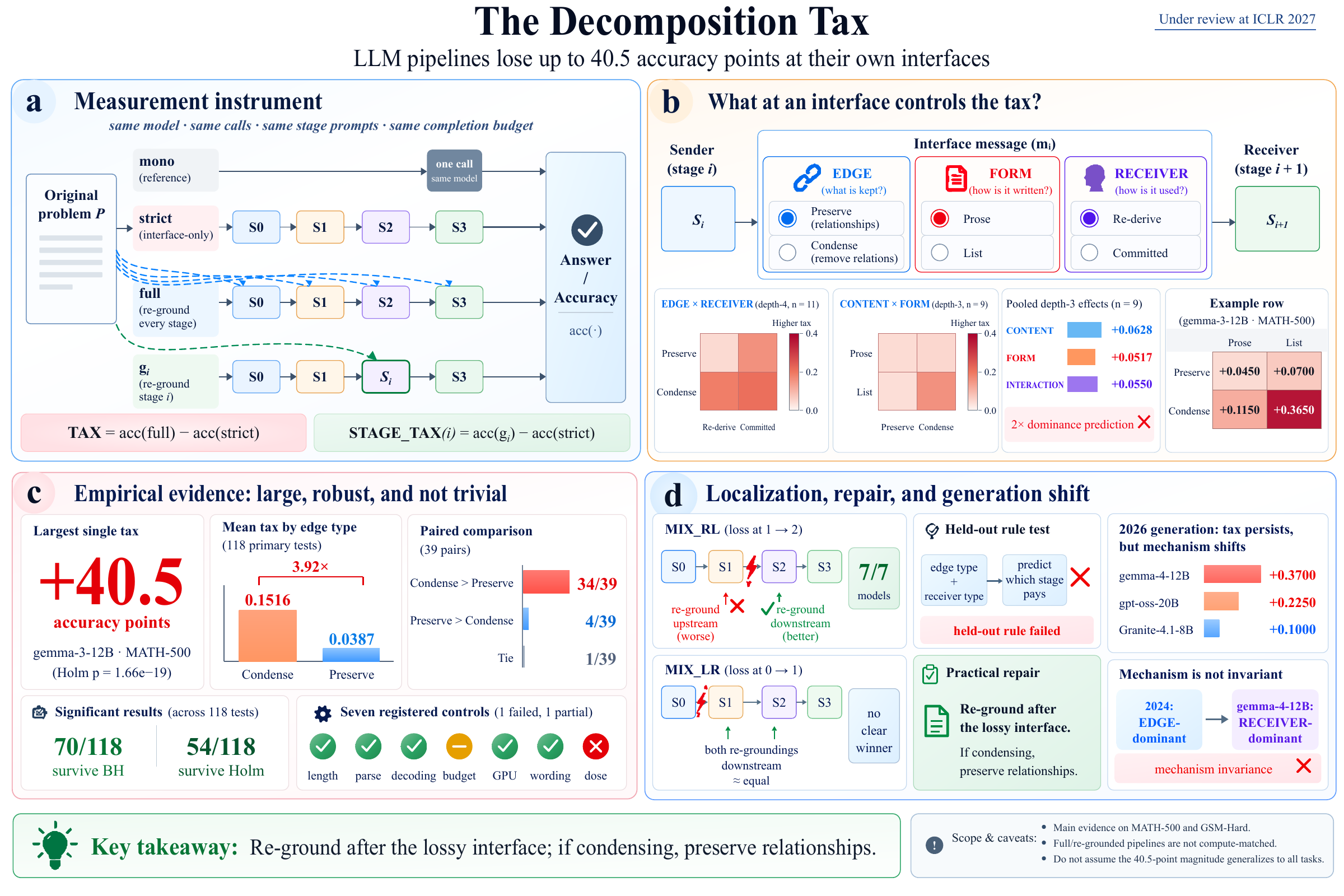}
\caption{\textbf{The decomposition tax at a glance.} \textbf{a} The measurement: the arms share model, calls, stage prompts and completion budget and differ only in which stages see the original problem. \textbf{b} What is varied at one interface, with the mean tax of each square's corners. \textbf{c} How large the tax is and the controls behind it. \textbf{d} Where re-grounding helps, the rule that failed its sealed test, and the 2026 cohort.}
\label{fig:fig0_author}
\end{figure}

A pipeline is a fixed sequence of stages, each a single LLM call with a
fixed prompt. Stage 0 reads the problem; every later stage reads only
the previous stage's output, and its prompt states that it does
\textbf{not} have the original problem. The final stage emits the answer
in a fixed format. Re-grounding prepends the original problem to a
stage's input under the header ``Original problem (for reference)'' and
leaves the stage prompt byte-identical. Three properties of one
interface vary, and EDGE names the first two together:

\begin{itemize}
\tightlist
\item
  \textbf{CONTENT}: what the sender's instruction asks it to pass on.
  \emph{Condensing}: ``the numerical quantities given and exactly what
  is asked''. \emph{Preserving}: the same, plus ``every relationship
  stated between them''.
\item
  \textbf{FORM}: how the sender writes. \emph{List}: ``a clean
  structured list''. \emph{Prose}: ``a plain-prose restatement in your
  own words''.
\item
  \textbf{EDGE}: the sender's whole instruction, content and form
  together. The depth-4 square and the mixed-order pipelines contrast a
  condensing list with a preserving prose restatement.
\item
  \textbf{RECEIVER}: what the next stage may do. \emph{Committed}:
  ``Output only the expression(s) to compute.'' \emph{Re-deriving}: the
  same stage told to ``reason step by step and then write the arithmetic
  expression(s)'' and to ``End with the expression(s) to compute.''
\end{itemize}

We name each square by its pipeline's depth, the number of stages. Its
four cells differ only in the strings quoted above, with every other
stage byte-identical; a build-time check enforces this. We call an
interface lossy when its sender condenses and restating when it
preserves. \texttt{MIX\_LR} puts a lossy interface first and a restating
one second; \texttt{MIX\_RL} reverses them, so its loss sits between
stages 1 and 2. We hand-wrote the headline pipelines. \texttt{A\_asset}
has four stages, and its first lists the quantities and the question.
\texttt{B\_asset} has three, and its first must ``restate, in your own
words, the given quantities and exactly what must be found''. The edge
classifier fixed in our registration, which labels an interface from its
prompt text alone (Appendix~D), counts that restatement as preserving.
The benchmarks are GSM-Hard \citep{gao2022pal} and MATH-500
\citep{hendrycks2021measur} at n = 200 items per cell. The 21
open-weight models, from 1B to 24B parameters, run locally at
temperature 0. OLMo-2-13B's short context window forces a smaller
completion budget, so the \texttt{A\_asset} counts cover the other 20
(Appendix~A, Appendix~C).

\textbf{Statistics.} Every cell reports a paired \textbf{McNemar} exact
test \citep{mcnemar1947note} and a paired bootstrap interval. We correct
each claim for multiplicity within its family (primary, structural,
control, sampled, exploratory) with both Benjamini-Hochberg FDR
\citep{benjamini1995contro} and Holm's family-wise adjustment
\citep{holm1979a}. Families follow from each cell's recorded
configuration, never from outcomes (Appendix~D). The primary family
holds the greedy full-versus-strict tests of \texttt{A\_asset},
\texttt{B\_asset} and the depth-3 square's condensing-list pipeline on
both benchmarks: 118 tests.

\section{3. Results}\label{results}

A named script regenerates every number below from the corpus of
\textbf{445 distinct cells} readable under Appendix~D's parse floor,
available from the corresponding author on request, and \texttt{code/claims\_trace.py} checks every
four-decimal value against this text. Two cohorts recur, named as in the
registration: the \textbf{2024 cohort} (Llama-3, Qwen2.5, Gemma 3,
Phi-4, OLMo-2, Ministral, Mistral-Small) and the \textbf{2026 cohort},
the three newest models we test (gemma-4-12B, gpt-oss-20B,
Granite-4.1-8B). The names label groups, not release years: Gemma 3 and
gpt-oss both carry 2025 references. The Qwen3 and R1-distilled models
belong to neither.

\subsection{3.1 The tax reaches 40.5 points (Figure
2)}\label{the-tax-reaches-40.5-points-figure-2}

\begin{figure}[t]
\centering
\includegraphics[width=1.00\linewidth]{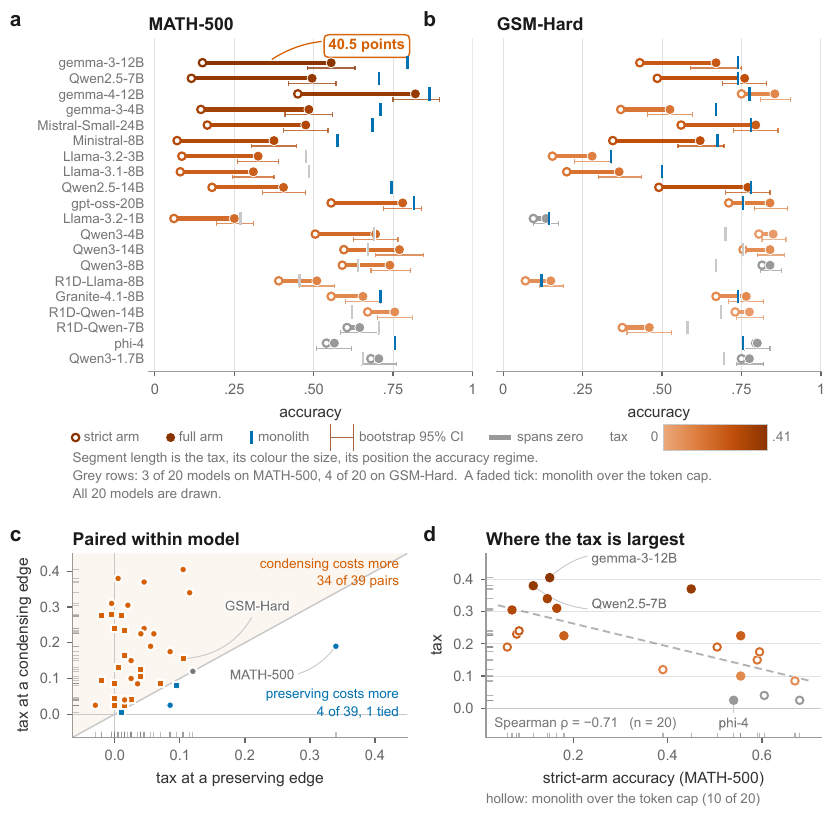}
\caption{\textbf{The tax is large across the corpus, and a condensing edge costs far more than a preserving one.} \textbf{a,b} Per-model tax on each benchmark, with bootstrap 95\% intervals; a faded monolith tick marks models whose monolith hits the token cap. \textbf{c} The same cells paired within model: points above the diagonal are models where condensing costs more than preserving. \textbf{d} Tax against strict-arm accuracy, part of which the tax contains by definition; hollow markers are monoliths over that same cap.}
\label{fig:fig1_tax}
\end{figure}

The largest tax in the primary family is \textbf{gemma-3-12B on
MATH-500, \texttt{A\_asset}: +0.4050, Holm-corrected p = 1.66e-19}
across all 118 primary tests. The tax is common: on \texttt{A\_asset}
its bootstrap interval excludes zero for \textbf{17 of 20} models on
MATH-500 and \textbf{16 of 20} on GSM-Hard, and \textbf{70 of 118}
primary-family tests survive Benjamini-Hochberg (\textbf{54} survive
Holm). Keeping one test per cell (this drops eight greedy re-runs) and
applying the registered parse gate on the strict arm leaves \textbf{52
of 94} tests surviving Benjamini-Hochberg and \textbf{43} Holm, with
this cell at p = 1.32e-19 (\texttt{out/headline\_family.txt}). Appendix~D applies the registered parse and empty-output gates in turn. The
second, no empty stage output, would exclude this cell: two of its 200
full-arm items carry an empty intermediate message from the second
stage. We keep it, because every final answer in that arm parses and two
items in 200 can move the tax by at most one point.

{\def\LTcaptype{none} %
\begin{center}
\small
\begin{tabular}{@{}lrrr@{}}
\toprule\noalign{}
edge & tests & survive BH & mean tax \\
\midrule\noalign{}

\textbf{CONDENSING} & 76 & \textbf{59} & \textbf{+0.1516} \\
\textbf{PRESERVING} & 42 & 11 & +0.0387\\
\bottomrule
\end{tabular}
\end{center}

}

A 3.92x ratio over the 118 primary tests above clears the 3x floor we
registered before extending \texttt{B\_asset}, the preserving pipeline,
to the full roster. Condensing costs more in 34 of 39 within-model pairs
of \texttt{A\_asset} and \texttt{B\_asset} (Figure 2c). At matched depth
the gap holds: the three-stage condensing-list pipeline averages
\textbf{+0.1362} over its primary cells against +0.0387 for the
three-stage \texttt{B\_asset} (\texttt{out/multcomp.txt}). In every
primary pipeline the condensing interface also writes a list and the
preserving one writes prose; Section 3.2 separates what the sender keeps
from how it writes.

\textbf{Re-grounding closes most of the gap to a single call.} Over the
322 cells where a single call beats the strict pipeline by more than
0.05, re-grounding recovers a median \textbf{63\%} of the gap, and 65\%
on \texttt{A\_asset}. On the headline cell, accuracy rises from 15.0\%
to 55.5\%; a single call reaches 79.5\% (Appendix~H discusses the single
call's token cap).

\textbf{The tax is the missing problem.} Registered controls address the
obvious alternatives (Appendix~D). A placebo on \texttt{A\_asset}, run
on GSM-Hard, replaces all but at most one of the problem's words with
\texttt{\_\_\_} while keeping its layout and word count. It carries at
least \textbf{60\%} of full re-grounding's extra tokens and recovers
\textbf{-0.0100 / -0.0050 / -0.0100} on its three models. Twenty sampled
seeds on Qwen2.5-7B average \textbf{+0.3922} against a greedy +0.3800 on
MATH-500, and no MATH-500 strict or full arm at the 4096-token corpus
budget parses below \textbf{0.955}. The same cell on a second GPU
architecture gives \textbf{+0.3850} against \textbf{+0.3800}, and
rewording the depth-4 square without mathematical vocabulary leaves its
condensing tax in place on 4 of 5 models (Section 3.2). A budget sweep
over other models, run in place of the registered headline-cell control,
moves the tax by at most \textbf{0.055} on MATH-500. Re-grounding half
of the problem recovers almost nothing, so recovery does not grow in
step with how much of the problem a stage sees. The registered
proportional prediction held on 0 of 3 models. Across cells the full arm
reads a median \textbf{1.17$\times$} the strict arm's tokens (Appendix~E).

\textbf{Takeaway.} The tax is large, common across model families and
concentrated at condensing interfaces. Re-grounding buys back the
problem's content; in the placebo, extra length without that content
bought nothing.

\subsection{3.2 One stage's instruction moves the tax from 4.5 to 36.5
points (Figure
3)}\label{one-stages-instruction-moves-the-tax-from-4.5-to-36.5-points-figure-3}

\begin{figure}[t]
\centering
\includegraphics[width=1.00\linewidth]{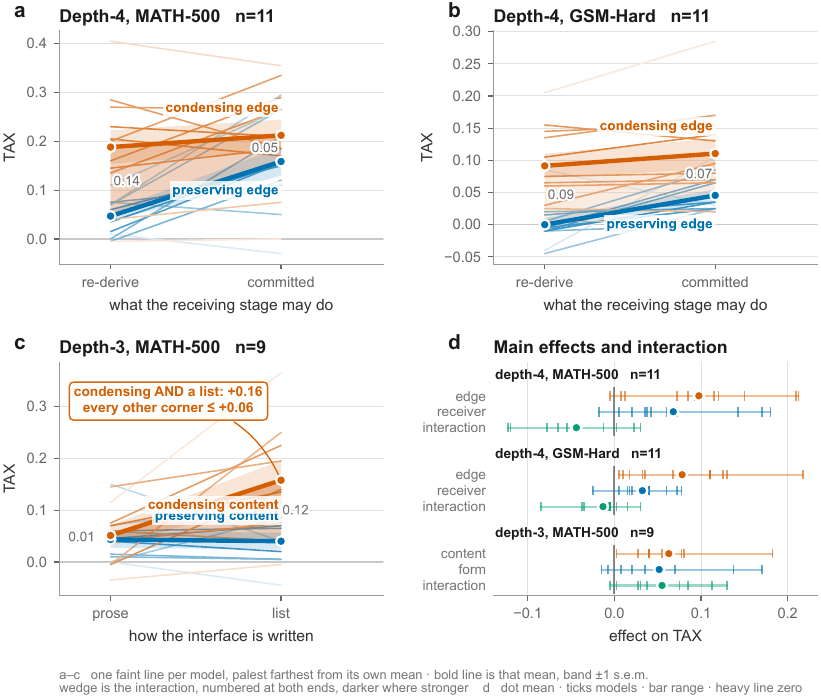}
\caption{\textbf{What the sender keeps, how it writes and what the receiver may do all move the tax.} \textbf{a,b} The depth-4 square on MATH-500 and GSM-Hard, across the receiver factor. \textbf{c} The depth-3 square on MATH-500, across output form; all four of its corners use mathematical wording (Section 3.2). In \textbf{a} to \textbf{c}, faint lines are models and bold lines their means. \textbf{d} Each square's main effects and interaction.}
\label{fig:fig2_mechanism}
\end{figure}

On gemma-3-12B, changing only what one stage's instruction asks it to
keep and how to write it moves the tax from \textbf{+0.0450} to
\textbf{+0.3650} inside the depth-3 square on MATH-500:

{\def\LTcaptype{none} %
\begin{center}
\small
\begin{tabular}{@{}lrr@{}}
\toprule\noalign{}
& prose & list \\
\midrule\noalign{}

\textbf{preserve} & +0.0450 & +0.0700 \\
\textbf{condense} & \textbf{+0.1150} & \textbf{+0.3650}\\
\bottomrule
\end{tabular}
\end{center}

}

Across the depth-3 square's nine models from seven organisations
(\texttt{out/square\_depth3\_4096.txt}): Mean \textbf{CONTENT +0.0628,
positive on 9 of 9}; Mean FORM +0.0517, positive on 7 of 9; mean
interaction \textbf{+0.0550}. On MATH-500, form matters about as much as
content: content leads by 1.22x over the nine models at 4096 tokens, and
form edges ahead once OLMo-2-13B joins at its own budget (Appendix~C).
On GSM-Hard, where we registered it, our prediction that content would
reach 0.08 and lead form by 2x held on at most 1 of 4 models, and form
stayed within 0.05 of zero on 3 of 4, as registered
(\texttt{out/amend10\_score.txt}). In the 2024 cohort the two compound,
so a condensing list costs more than condensing alone and a list alone
add up to. The depth-4 square crosses the edge, here a condensing list
against a preserving prose restatement, with the receiver. Across its 11
models a condensing edge raises the tax on 11 of 11 on GSM-Hard and 10
of 11 on MATH-500, and a committed receiver raises it on 10 of 11 on
each (Figure 3a,b; \texttt{out/square\_depth4.txt}).

\textbf{The relationships clause lowers the tax.} When the output stays
a list, telling the stage to keep ``every relationship stated between
them'' lowers the tax on \textbf{9 of 9 models on MATH-500} and
\textbf{3 of 4 on GSM-Hard}. On gemma-3-12B it falls from 36.5 points to
7.0; we had registered a target of 5 of 7 models, before the square grew
to nine. The one GSM-Hard exception, Qwen2.5-7B, pays +0.0200 with and
without the clause.

\textbf{Answer extraction does not produce the form effect.} Much of the
reported prompt sensitivity is rigid answer matching
\citep{hua2025flaw}. Here \texttt{parse\_rate} reads \textbf{1.000} in
157 of 172 MATH-500 cells at the corpus budget, counting the lower of
the strict and full arms. It reads \textbf{0.9550} in the worst, or 9
unparseable items in two hundred, against a 25-point list-versus-prose
gap on gemma-3-12B.

\textbf{The depth-4 tax survives rewording.} Every corner above asks for
``the numerical quantities given''. On the depth-4 square's condensing,
committed corner, rewording every stage without mathematical vocabulary
leaves gemma-3-12B paying +0.4500 against +0.3550 with the vocabulary
(\texttt{out/amend9\_score.txt}). At depth 3 the same rewording removes
between 0.1150 and 0.3500 of the condensing list's MATH-500 tax on each
of the seven models with a tax (\texttt{out/oneword\_effect.txt}), so
the depth-3 square measures content and form inside a mathematical
wording.

\textbf{Takeaway.} A builder sets much of the tax in one stage's prompt.
If a stage must list the numerical quantities, tell it to keep the
relationships; at a lossy edge, prose also costs less than a list.

\subsection{3.3 With one lossy interface, the repair belongs after it
(Figure
4)}\label{with-one-lossy-interface-the-repair-belongs-after-it-figure-4}

\begin{figure}[t]
\centering
\includegraphics[width=1.00\linewidth]{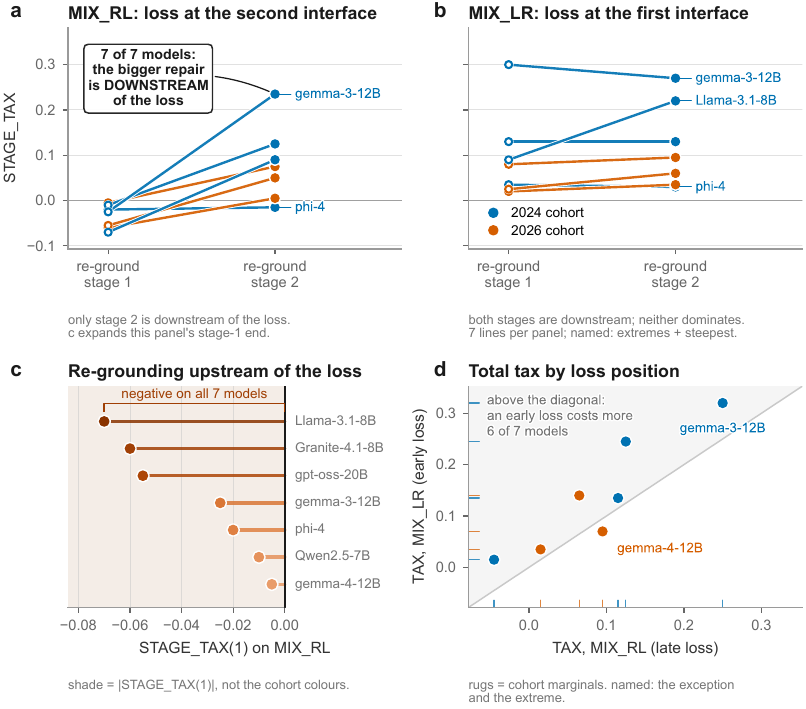}
\caption{\textbf{Which stage is re-grounded decides the recovery, and the intuitive choice is the wrong one.} All panels MATH-500. \textbf{a} MIX\_RL, the loss at the second interface: re-grounding stage 2 recovers more than stage 1 on 7 of 7 models. \textbf{b} MIX\_LR, the loss at the first interface: both stages are downstream and neither dominates. \textbf{c} Stage 1 re-grounded alone on MIX\_RL, upstream of the loss: negative on all 7 models, worse than no repair. \textbf{d} Total tax with the loss early against late; the registered order test is in Appendix~F.}
\label{fig:fig3_mix}
\end{figure}

When a budget lets only one stage see the problem again, the choice of
stage decides the recovery. The intuitive choice is the earliest stage,
since everything downstream inherits from it. \texttt{MIX\_RL}, whose
loss sits between stages 1 and 2, tests that choice against the stage
after the loss, and the registration named this test the decisive one
(\texttt{out/amend11\_score.txt}, \texttt{out/amend17\_score.txt}).

\textbf{Re-grounding the stage after the loss beats re-grounding the
stage before it on 7 of 7 models} on both benchmarks. That is 4 of 4 in
the 2024 cohort, and 3 of 3 in the 2026 cohort, for which we registered
the MATH-500 test before its cells ran. On MATH-500 re-grounding the
stage before the loss does worse than no repair on 7 of 7 (by 0.005 to
0.070; p = 0.016, two-sided sign test), against 3 of 7 on GSM-Hard. On
gemma-3-12B, stage 1 alone recovers -0.0250; stage 2 alone recovers
+0.2350 of the +0.2500 that re-grounding every stage recovers. In
\texttt{MIX\_LR}, where both re-grounded stages sit after the loss, the
two help about equally, and our registered prediction that stage 1 would
help more failed. By its registered reading, which of two stages after
the loss helps more is then set by the stage's own prompt or its
position, not by which interface precedes it. Stage 2 runs the same
\texttt{solve} prompt in both pipelines (Appendix~F). Stage 1 is not
the \texttt{solve} stage in either pipeline, and on MATH-500
re-grounding it helps on 7 of 7 models where it sits after the loss and
hurts on 7 of 7 where it sits before it.

\textbf{A sealed test rejects a stronger rule.} We also registered a
rule that predicts which stage pays from the interface type and the
receiver type: a committed receiver pays and a re-deriving one does not.
Before either of two held-out pipelines ran, we sealed seven predictions
and the decision rule. Of the four predictions on which this rule and a
rival using the interface type alone disagree, the rule lost three and
the fourth was void, so we count it refuted. The rival predicted that a
re-deriving stage behind a condensing interface still pays; this held on
4 of 4 models (Appendix~F). The repair direction comes from other
pipelines and stands, and we measure incidence one stage at a time.

\textbf{Takeaway.} If a budget lets one stage see the original problem
again, give it to the stage after the lossy interface. On MATH-500 the
intuitive choice, the stage before it, is worse than no repair on 7 of 7
models.

\subsection{3.4 Newer models still pay, and the repair still holds
(Figure
5)}\label{newer-models-still-pay-and-the-repair-still-holds-figure-5}

\begin{figure}[t]
\centering
\includegraphics[width=1.00\linewidth]{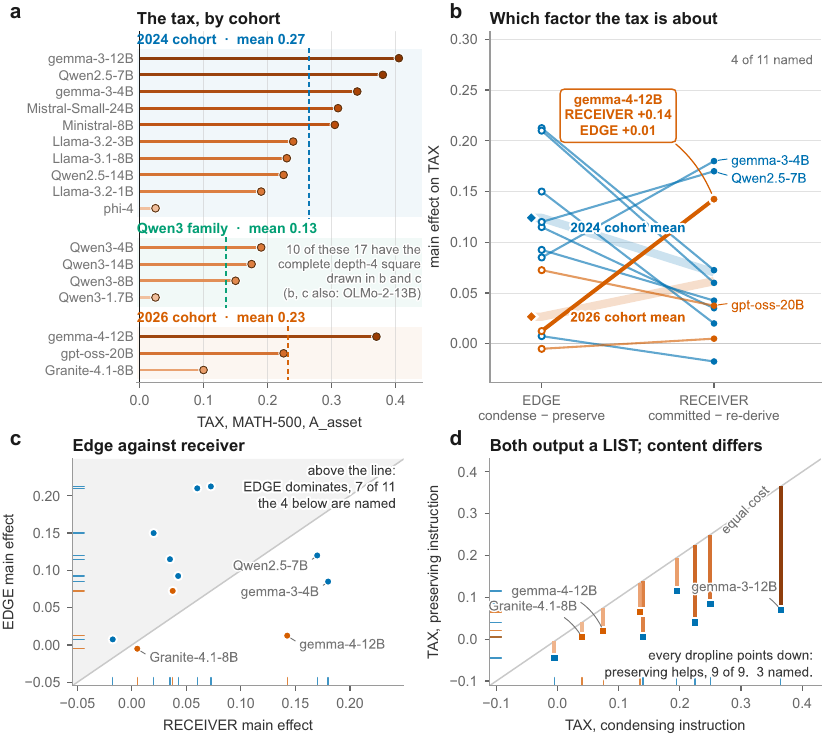}
\caption{\textbf{The tax survives into the 2026 cohort; on MATH-500 its make-up shifts.} \textbf{a} Tax on MATH-500, A\_asset, per model, in the cohorts of Section 3; vertical rules mark cohort means. \textbf{b} Each model's edge and receiver main effects on the depth-4 square, drawn as a slope; thick lines are cohort means. \textbf{c} Edge against receiver effect, with the diagonal. \textbf{d} The depth-3 square with list output: tax under the condensing against the preserving instruction, both in mathematical wording.}
\label{fig:fig4_generation}
\end{figure}

We registered the outcome that would have changed this paper's title:
that the 2026 cohort no longer pays. All three models still pay and
survive Benjamini-Hochberg on MATH-500, \texttt{A\_asset}
(\texttt{out/amend12\_score.txt}):

{\def\LTcaptype{none} %
\begin{center}
\small
\begin{tabular}{@{}lrrrrrr@{}}
\toprule\noalign{}
model
 & tax
 & Holm p
 & BH q
 & monolith
 & strict
 & monolith cap hits
 \\
\midrule\noalign{}
\textbf{gemma-4-12B} & \textbf{+0.3700} & 8.68e-17 & 1.8e-17 & 0.865 &
0.450 & 3 / 200 \\
gpt-oss-20B & +0.2250 & 7.09e-11 & 5.37e-12 & 0.815 & 0.555 & 16 /
200 \\
Granite-4.1-8B & +0.1000 & \textbf{0.0526} & 0.00176 & 0.710 & 0.555 & 6
/ 200\\
\bottomrule
\end{tabular}
\end{center}

}

We had also registered that the three 2026 models would pay, on average,
at least 0.10 less than gemma-3-12B, Qwen2.5-7B and Llama-3.1-8B. At a
matched 4096-token budget their mean tax is \textbf{+0.2317} against
\textbf{+0.3383}, a drop of \textbf{0.1067} that clears the registered
0.10; Figure 5a's 2024 mean pools the ten 2024-cohort models run at 4096
tokens. Appendix~G gives the budget rule and the registered readings
of the drop.

On MATH-500, the make-up of the tax shifts. In the 2024 cohort the
depth-4 edge factor averages \textbf{+0.1241} against \textbf{+0.0703}
for the receiver and leads on 6 of 8 models
(\texttt{out/square\_depth4.txt}); gemma-4-12B reverses the order, at
\textbf{+0.1425} for the receiver and \textbf{+0.0125} for the edge. We
had registered this structure as invariant across cohorts, and both
registered tests of that invariance fail, one on the depth-4 edge effect
and one on the depth-3 content-and-form interaction (Appendix~G).

Before running the repair test on this cohort, and after those tests had
failed, we registered how a pass would read: the make-up is
generation-specific and the repair is not
(\texttt{out/amend17\_score.txt}). The repair passed on all three models
(Section 3.3), and they are three of the nine on which the relationships
clause lowers the tax on MATH-500 (Section 3.2, Figure 5d).

\begin{quote}
\textbf{The tax and the repair persist into the 2026 cohort. On MATH-500
the 2024 cohort pays mostly through the edge; gemma-4-12B pays mostly
through whether the receiver may re-derive.}
\end{quote}

\section{4. Related work}\label{related-work}

\textbf{Decomposition and its cost.} Least-to-most prompting
\citep{zhou2022leastt} and Decomposed Prompting \citep{khot2022decomp}
split a problem across calls, and PAL \citep{gao2022pal} hands the
solving step to an interpreter. Each measures what its design buys, and
recent work adds error recovery to it
\citep{hernndezgutirrez2025recurs}. \citet{radhakrishnan2023questio}
measure what an isolated context costs, but their arms change the call
structure and what each call sees together, and none is both decomposed
and re-grounded. Our arms differ in what each stage sees alone. DnDScore
\citep{wanner2025dndsco} checks decomposed subclaims against their
source context. This is our \texttt{full} arm used as a method; we use
it as an instrument.

\textbf{Failures at the interface.} MAST \citep{cemri2025why} catalogues
multi-agent failures, MAS-FIRE \citep{jia2026masfir} injects faults, and
AgentAsk \citep{lin2026agentask} repairs erroneous messages at the
interface with a clarifying question. Each studies an error at the
interface. The tax is what a working interface costs with no fault
injected, and our repair adds no call. In software engineering, a stage
that reads an earlier stage's output also does worse than one that works
independently \citep{konstantinou2026on}.

\textbf{Format, context and compute.} \citet{sclar2023quanti} vary a
whole prompt's format, and \emph{Lost in the Middle} \citep{liu2024lost}
varies where information sits. We vary what one interface carries with
every other byte fixed, and a placebo separates missing content from
missing length. Test-time compute work varies the budget
\citep{snell2024scalin,wu2024infere,wang2025genera}; we fix the
completion cap and vary only what each stage can see. The decomposition
tax is what these lines leave unmeasured: one interface, with the rest
of the pipeline fixed. Appendix I gives the full account.

\section{5. Conclusion}\label{conclusion}

Builders design a pipeline one stage at a time, and its bill arrives at
the interfaces between stages. With model, stages, stage prompts and
completion budget fixed, the tax reaches \textbf{40.5 accuracy points},
and \textbf{70 of 118} primary-family tests survive Benjamini-Hochberg
correction. The tax is the missing problem: on GSM-Hard, a placebo
carrying at least 60\% of the extra tokens recovers nothing. One stage's
instruction moves it from 4.5 to 36.5 points, and with one lossy
interface the repair belongs after it in both cohorts. The prescription
has two parts: re-ground the stage after the lossy interface, and if a
stage must list the quantities, tell it to keep the relationships. A
sealed test refuted a stronger registered rule that predicted which
stage pays from the interface and receiver types; the tax and the repair
rest on other pipelines and stand.

\begin{center}\rule{0.5\linewidth}{0.5pt}\end{center}

\label{endofmain}
\section{Reproducibility statement}\label{reproducibility-statement}

Every prediction this paper scores was registered in a dated file
\textbf{before} it was scored; Amendment 12 was written after some of
its cells had landed, and its title says so. The file and its
twenty-four amendments are available with the code from the corresponding author on request, numbered to 25
because number 23 was drafted and not adopted, including the amendments
that record a prediction failing, the one that withdraws the original
confirmatory family, and the ones that correct an earlier amendment.

No measured number in this paper is typed by hand; each is emitted by a
named script from result JSONs. \textbf{19 checks} run over the
manuscript and the corpus on every build, and \textbf{17 of them ship a
committed negative test that shows the check failing on the defect it
exists to catch}. A check never seen to fail is not evidence. The two
without one report inventory rather than pass or fail a claim, and are
marked as such rather than counted as if they were covered. The full
list is in \texttt{code/gates.py}, which is also what runs them; three
carry most of the weight:

\begin{itemize}
\tightlist
\item
  \texttt{claims\_trace.py}: every four-decimal value in the text must
  trace to a script output or a claimable field of a result file. Values
  that are historical, such as a figure quoted inside its own correction
  note, are declared individually with a pinned maximum count, and a
  declaration wider than the value needs is itself a failure.
\item
  \texttt{cite\_trace.py}: every citation named in the prose must
  resolve to a verified bibliography row, and the check fails if it
  finds fewer citations than a pinned floor, so it cannot pass by not
  looking.
\item
  \texttt{consistency.py}: a label such as ``the edge effect'' may not
  take two different values in the manuscript.
\end{itemize}

Results land in the corpus only through \texttt{land\_results.py}, which
refuses any file whose arms carry no \texttt{n}, whose comparison has no
bootstrap interval, or whose reported delta does not equal
\texttt{acc(full)\ -\ acc(strict)} recomputed from the arms. Twelve
partial files were refused during this campaign; each was a cell
interrupted mid-write.

\textbf{Code and data.} The code and per-run logs are available from the
corresponding author on request: the pipeline
definitions, the campaign driver, every analysis and figure script, the
registration file with its amendments, and all 1,025 result files. Each
cell's file carries the manifest it ran under (model, shape, task, seed,
sampling parameters, token budget, item-set checksum and, for cells run
after the field was added, the GPU), so any pooling assumption in this
paper can be re-checked against the data rather than taken on trust. The
figure scripts read only those files and reproduce the figures in this
paper pixel for pixel.

\textbf{Compute.} Cells ran on a shared academic SLURM cluster across
four GPU types: RTX PRO 5000 Blackwell, RTX 4500 Ada, RTX A6000 and RTX
A4500. Appendix~D reports the one registered cross-architecture
control, on Ada; it does not cover the two Ampere cards, which ran 10 of
the result files.

\section{AI use statement}\label{ai-use-statement}

Generative AI tools were used solely to assist with language editing and
improve the clarity and readability of the manuscript. All scientific
content, analyses, results, interpretations, and conclusions were
developed and verified by the authors, who take full responsibility for
the content of the paper.

\section{Ethics and broader impact}\label{ethics-and-broader-impact}

This work uses public benchmarks and released open-weight models; no
human subjects, no personal data, and no new model was trained. The
models are run locally for evaluation only.

The practical risk in a result like this is over-application. The
prescriptions in this paper are measured on two mathematical-reasoning
benchmarks with a fixed pipeline family. We do not know that they
transfer to retrieval, planning, code, or tool-use pipelines, and what
evidence we have points the other way on magnitude: on the one
non-mathematical task we could measure cleanly the tax is
\textbf{smaller on 4 of 4 paired comparisons over 2 models}, which is a
direction and not a result (Appendix~H, limitation 8). A reader who
prefers prose to a list at a lossy edge in a setting where the list is
machine-parsed downstream would be extrapolating past our evidence, and
one who assumes the 40.5-point figure travels to a non-mathematical
pipeline would be extrapolating past it in the other direction.

The result is more likely to reduce compute than increase it: it says a
cheaper interface choice recovers accuracy that would otherwise be
chased with more stages or more sampling.

\bibliography{refs}
\newpage
\appendix
\subsection{A Data, models, serving}\label{data-models-serving}

GSM-Hard \citep{gao2022pal}, the adversarial rewrite of GSM8K
\citep{cobbe2021traini}, and MATH-500 \citep{hendrycks2021measur}, n =
200 items per cell, temperature 0 unless stated.

The roster is 21 open-weight instruction-tuned models from nine
organisations, spanning two generations: Qwen2.5 \citep{qwen2024qwen}
and Qwen3 \citep{yang2025qwen}; Llama 3.1 and 3.2
\citep{grattafiori2024the}; Gemma 3 \citep{gemma2025gemma}; Phi-4
\citep{abdin2024phi}; OLMo-2 \citep{olmo2024}; the DeepSeek-R1
distillations \citep{deepseekai2025deepse}; gpt-oss
\citep{openai2025gptoss}; and the Mistral, Ministral and Granite
families. The \texttt{mono} reference arm is a single chain-of-thought
call \citep{wei2022chaino}; no self-consistency \citep{wang2022selfco}
or other sampling aggregation is used in any arm, since the comparison
is between two pipelines at a matched budget rather than between
decoding strategies.

Models are served locally with vLLM \citep{kwon2023effici}; both arms of
a cell run in the same job on the same GPUs, so no paired comparison
spans hardware. Every result file records the model snapshot hash, the
item-set md5, the code md5s, the effective token budget, and per-stage
diagnostics.

\subsection{B Two diagnostics that decide whether a cell means
anything}\label{two-diagnostics-that-decide-whether-a-cell-means-anything}

Reported beside every tax, because both have invalidated a result in
this corpus:

\begin{itemize}
\tightlist
\item
  \textbf{\texttt{parse\_rate}}: the share of a stage's outputs the
  answer extractor can read. A cell whose strict arm parses 4.5\% of its
  outputs is measuring a formatting collapse, not reasoning. One
  headline of ours was retracted on exactly this ground.
\item
  \textbf{\texttt{mono.cap\_total}}: how often the reference call hits
  the completion cap. For a long-reasoning model this can reach 34.5\%
  at a 4096-token budget, which makes the \emph{reference point}
  incomparable even though it leaves the tax itself alone.
\end{itemize}

\begin{center}\rule{0.5\linewidth}{0.5pt}\end{center}

\subsection{C What each claim is measured
over}\label{what-each-claim-is-measured-over}

The corpus holds \textbf{21 open-weight models}, but no single cell
shape covers all of them, and a claim is only as wide as the cells that
carry it. The headline \texttt{A\_asset} design covers \textbf{20 models
on MATH-500 and 20 on GSM-Hard}, of which \textbf{17} and \textbf{16}
carry the claim; the rest have bootstrap intervals spanning zero and are
drawn grey in Figure 2. The 2x2 designs are narrower, because a square
needs all four corners and a model with three of them contributes to no
main effect and no interaction: the depth-4 square covers \textbf{11}
models on each benchmark, and the depth-3 square \textbf{9} on MATH-500
and 4 on GSM-Hard. Every figure states its own n, and the pooled means
in Section 3 are over the models present in that comparison rather than
over the roster.

Every count here is recomputed from the corpus by
\texttt{code/roster\_coverage.py} under the same admissibility rules the
figures draw with, and checked against this prose at build time.

The gaps are not a sample. At the corpus budget exactly \textbf{one}
model is absent, on both benchmarks: \textbf{OLMo-2-13B}, whose
4096-token context window means it runs only at \texttt{max\_tokens}
1024 and is therefore not budget-matched to the rest, an exclusion by
design rather than a cell that has not run. Figure 2 and the coverage
counts pin the budget and so exclude it. The depth-4 square includes it
at its own budget, because a square compares corners within one model,
and so does multiplicity, because it depends on how many tests ran, and
this model's ran at the budget it was meant to have. The depth-3 square
does not include it: it is pinned at 4096, the budget Amendment 16
re-ran it at, although OLMo-2-13B has all four depth-3 corners at 1024.
With them the MATH-500 square has ten models, and FORM
(\textbf{+0.0708}) edges CONTENT (\textbf{+0.0667})
(\texttt{out/square\_depth3\_math\_all.txt}). Appendix~H states what that
leaves open.

\textbf{The arms.} For a pipeline with stages 0..k$-$1 we run:

{\def\LTcaptype{none} %
\begin{center}
\small
\begin{tabular}{@{}l>{\raggedright\arraybackslash}p{(\linewidth - 2\tabcolsep) * \real{0.5000}}@{}}
\toprule\noalign{}
arm
 & what each stage sees
 \\
\midrule\noalign{}
\texttt{mono} & one call, chain-of-thought, the whole problem (a
reference point) \\
\texttt{strict} & stage \emph{i} sees stage \emph{i$-$1}'s output only \\
\texttt{full} & stage \emph{i} additionally sees the original problem \\
\texttt{g\_i} & \emph{only} stage \emph{i} additionally sees the
original problem\\
\bottomrule
\end{tabular}
\end{center}

}

\textbf{What one cell resolves.} At n = 200 the median bootstrap 95\%
interval on a per-cell tax is \textbf{0.1100} wide, so a single cell
resolves about \textbf{+-0.0550} (\texttt{out/power\_audit.txt}, over
444 cells; 265 of them, 60\%, have intervals excluding zero). The
depth-3 square's CONTENT (+0.0628) clears that line; its FORM (+0.0517)
and interaction (+0.0550) are nine-model means whose sign most of the
nine models share (FORM is positive on 7 of 9), and gemma-3-12B resolves
the interaction on its own, at \textbf{+0.1125}.

\subsection{D Seven controls, and what each
registered}\label{seven-controls-and-what-each-registered}

\textbf{The five families.} \texttt{out/multcomp.txt}, from
\texttt{code/multcomp.py}, corrects each family on its own, so a
headline is never pooled with controls built to test something else:

{\def\LTcaptype{none} %
\begin{center}
\small
\begin{tabular}{@{}l>{\raggedright\arraybackslash}p{(\linewidth - 8\tabcolsep) * \real{0.5631}}rrr@{}}
\toprule\noalign{}
family
 & what it is
 & m
 & survive BH
 & survive Holm
 \\
\midrule\noalign{}
\textbf{PRIMARY} & the pre-registered tax on the two real benchmarks &
118 & \textbf{70} & \textbf{54} \\
STRUCTURE & the 2x2 designs at matched depth & 98 & 54 & 44 \\
CONTROL & placebo, dose-response, wording, hardware; only the placebo is
null by design & 63 & 27 & 23 \\
SEEDREP & sampled arms at T = 0.7 & 48 & \textbf{48} & \textbf{48} \\
EXPLORATORY & synthetic tasks and held-out shapes & 117 & 51 & 29\\
\bottomrule
\end{tabular}
\end{center}

}

Families are assigned from manifest fields alone and never from a cell's
outcome. The partition itself is not in the original registration; the
paragraph on the registration below gives its provenance. The 445th
cell, a pre-rename label re-run since, enters no claim.

\textbf{The seven controls.}

{\def\LTcaptype{none} %
\begin{center}
\small
\begin{tabular}{@{}>{\raggedright\arraybackslash}p{(\linewidth - 4\tabcolsep) * \real{0.2791}}>{\raggedright\arraybackslash}p{(\linewidth - 4\tabcolsep) * \real{0.1628}}>{\raggedright\arraybackslash}p{(\linewidth - 4\tabcolsep) * \real{0.5581}}@{}}
\toprule\noalign{}
control
 & prediction
 & result
 \\
\midrule\noalign{}
\textbf{Placebo} -- \texttt{frac\ =\ 0.01}, the share of words left
visible: all but at most one word masked to \texttt{\_\_\_}, structure
and word count kept & recovers nothing & \textbf{-0.0100 / -0.0050 /
-0.0100}, 3 of 3 inside +-0.05. It carries \textbf{60\%} of the extra
tokens full re-grounding carries on gemma-3-12B (+253 against +421),
61\% on Qwen2.5-7B (+139 against +228) and more than all of them on
Llama-3.1-8B (+395 against +364), and none of the benefit;
\texttt{out/placebo\_tokens.txt} \\
\textbf{Dose-response} & at frac 0.50 the tax is at least 30\% of the
full tax & \textbf{failed, 0 of 3}: flat through frac \textless= 0.5;
\textbf{frac 0.75 already recovers 40\%, 31\% and 15\%} of the total,
the rest between 0.75 and 1.00; Llama-3.1-8B's frac = 1.0 point is its
4096-token \texttt{A\_asset} cell (+0.1650), which Amendment 8 names \\
\textbf{Parse integrity} & not a formatting collapse & Across
\textbf{306} mathematical cells the strict arm's \texttt{parse\_rate}
averages \textbf{0.0026} below the full arm's -- the tax is not answers
the extractor could not read. at the corpus budget \texttt{parse\_rate}
is \textbf{1.000} in \textbf{157 of 172} MATH-500 cells and never below
\textbf{0.955}; the GSM-Hard floor is \textbf{0.735};
\texttt{out/parse\_recovery.txt}, \texttt{out/parse\_integrity.txt} \\
\textbf{Sampling} -- 20 seeds per task on Qwen2.5-7B's \texttt{A\_asset}
cell & not an artefact of greedy decoding & seeded mean \textbf{+0.3922}
against greedy +0.3800 on MATH-500 and \textbf{+0.2435} against +0.2750
on GSM-Hard, seed sd \textbf{0.0234} and \textbf{0.0203}; 48 of 48
sampled cells survive both corrections \\
\textbf{Token budget} -- a two- to fourfold sweep & not an artefact of
the completion cap & the registered control on the two headline cells
was not run; the sweep that was, over Qwen3, Llama-3.1, Qwen2.5 and
R1-distill, is stable to \textbf{0.055} on MATH-500 and 0.075 on
GSM-Hard, where two Qwen3 cells, 4B and 8B, cross zero at 16384 tokens
as their parse rates fall. The two registered predictions that a larger
cap would bring Qwen3's monolith below 10 truncations in 200 both failed
(T36; T41, 2 of 4 models at 16384 tokens), so Qwen3's single-call
accuracies are lower bounds (Appendix~H);
\texttt{out/budget\_sweep.txt} \\
\textbf{Hardware} -- the same cell on a different GPU architecture & not
an artefact of one card & tax \textbf{+0.3850} on Ada \texttt{sm\_89}
against \textbf{+0.3800} on Blackwell; no control covers the Ampere
cards; \texttt{out/t44\_score.txt} \\
\textbf{Wording} -- every stage rewritten without mathematical
vocabulary & the misfit account (T15): the four-stage tax falls inside
+-0.05 & \textbf{the misfit account fails, 1 of 5 in band}: at depth 4
the rewording removes none of the tax (gemma-3-12B \textbf{+0.4500}
against +0.3550), while at depth 3 it removes most of the MATH-500 tax
(Section 3.2). T16 passes only through GSM-Hard cells whose strict parse
falls to 0.735-0.965 and so supports nothing; T17, phi-4 unchanged,
passes; \texttt{out/amend9\_score.txt}\\
\bottomrule
\end{tabular}
\end{center}

}

\textbf{Does the headline depend on which cells count?} The family above
counts eight T = 0 re-runs of cells it already holds, and it does not
apply the registration's quality gates. Two are applied in turn below:
an answer parse rate of at least 0.99, which the registration sets for
every arm and we apply to the strict arm, and no empty stage output. A
third asks for at least 180 distinct messages at every stage. We read it
over the intermediate stages, because a final answer repeats across
items that share it: the headline cell's last stage carries 146 distinct
answers in the strict arm and 136 in the full arm. Read that way, it
would remove one of the 84 tests that pass the other rules, OLMo-2-13B's
\texttt{A\_asset} MATH-500 cell, at 168. Applying the first two in turn
(\texttt{out/headline\_family.txt}):

{\def\LTcaptype{none} %
\begin{center}
\small
\begin{tabular}{@{}>{\raggedright\arraybackslash}p{(\linewidth - 8\tabcolsep) * \real{0.1579}}rrrr@{}}
\toprule\noalign{}
rule, applied cumulatively
 & m
 & survive BH
 & survive Holm
 & headline cell, Holm p
 \\
\midrule\noalign{}
the primary family as run & 118 & 70 & 54 & 1.66e-19 \\
one test per cell, earliest run kept & 110 & 65 & 49 & 1.55e-19 \\
and strict-arm parse rate at least 0.99 & 94 & 52 & 43 & 1.32e-19 \\
and no empty stage output in either arm & 84 & 43 & 37 & excluded\\
\bottomrule
\end{tabular}
\end{center}

}

No rule for what counts as one test was registered, so the first row is
the family of record. We do not apply the last rule as a quarantine. It
was registered against a thinking-model defect that emptied 21 to 28 of
40 stage outputs; the headline cell fails it on 2 empty outputs in 200
items, both in the second stage of the re-grounded arm, which also
records two token-cap hits there. Every final answer in that arm parses,
and two items in 200 can move the tax by at most one point. The row is
shown so that a reader who would apply the rule can.
\texttt{out/multcomp.txt} also prints a robustness line per family at a
looser parse cut of 0.90, chosen after the fact: 2 of the 118 primary
tests fall below it, STRUCTURE has none, and the families that do are
CONTROL, at \textbf{4 of 63}, and EXPLORATORY, at \textbf{1 of 117}: the
wording arm and the synthetic tasks, where an answer extractor built for
two mathematics benchmarks fails hardest.

The control family behaves like one, with \textbf{27 of 63} surviving
BH. The arms that survive most strongly are ones expected to be
non-zero: the four-stage wording twins, which carry the tax without
mathematical vocabulary, and the hardware control, the headline-shape
tax re-run on a second GPU. The \texttt{frac\ =\ 0.75} cell, three
quarters of the problem rather than a placebo, survives at q = 0.0105.
The wording twins' largest cells, on GSM-Hard, are also among the
family's worst-parsing: gemma-3-12B pays \textbf{+0.6950} at a strict
parse of \textbf{0.910}, so part of that tax is answers the extractor
could not read, and those cells support nothing here.

\textbf{What the registration was written for, and where the family
partition came from.} The registration was written before any
local-model call for a different primary claim: a classifier, Edge-Typed
Grounding, that reads a pipeline's prompts and re-grounds only the
stages it selects, with the tax contrast as its negative control
(registration \S1). Its confirmatory family of seven predictions, P1 to
P5b (registration 5.4), was never scored and was withdrawn by Amendment
24 on 2026-09-22, a month after the campaign ended. Of those
predictions, P1, P2a, P2b, P5a and P5b are contradicted by data reported
here, and P4a is disfavoured. P1 predicted that the preserving shapes
carry no tax, with an interval inside +-0.02; they carry up to
\textbf{+0.3400} (\texttt{B\_asset}, Qwen3-4B, MATH-500), and at n =
200, where the median interval is 0.1100 wide (Appendix~C), an interval
that narrow is rare. P2a and P2b predicted that a stage pays only where
a condensing interface feeds a committed receiver, and pays more there
than where it feeds a re-deriving one; in \texttt{HO1\_cross} the
re-deriving stage pays more on all four models (Appendix~F). P5a and
P5b predicted a tax inside +-0.02 at \texttt{HO1\_cross}'s stage 3 and
across \texttt{HO2\_formal}; Appendix~F reports 0 of 4 models inside
that band for each. P4a predicted that the classifier-guided pipeline
beats a single call; that arm is not in this paper, and the fully
re-grounded pipeline, which re-grounds every stage the classifier could
select, wins 17 of 104 \texttt{A\_asset} cells (Limitation 1). The
sealed held-out predictions S1 to S7 (\S6) overlap them: S1 to S4 restate
P5a and P5b. Appendix~F scores all seven. Section 10 of the
registration lists framings it does not license, ``the decomposition
tax, measured'' among them; that entry recorded a positioning choice
made before the campaign, and this paper's framing departs from it. The
per-cell partition used here was never registered during the campaign:
only PRIMARY and SEEDREP are named in amendments, on 2026-08-20 and
2026-08-21, after 339 and 902 of the 1,025 result files had been
written, and the partition itself is first written down in Amendment 24,
after the campaign. What is assigned from the registration is
membership, since a cell's family follows from its manifest fields,
model, shape, task, temperature, re-grounding fraction and control flag,
and never from how it came out. What was arrived at during the campaign
is the partition those fields map onto. The individual controls in the
table above are each registered before they ran; the family they are
corrected within is not.

\textbf{Cells that are not measurements do not enter a family at all.} A
strict arm parsing below \textbf{0.50} is mostly the answer extractor
failing, so the gap the full arm appears to recover is mostly parse
recovery rather than reasoning. \textbf{22 further cells}, outside the
445, are excluded on that rule and named in \texttt{out/multcomp.txt}.
Every one is on a synthetic task or on MMLU; \textbf{not one is a
MATH-500 or GSM-Hard cell}, so the rule cannot reach the headline, and
PRIMARY is unchanged at 118 tests with 70 surviving BH either way.

\subsection{E Is the tax just more test-time compute? (post-hoc
audit)}\label{b-is-the-tax-just-more-test-time-compute-post-hoc-audit}

The full arm re-injects the original problem at every stage, so it
necessarily reads more tokens than the strict arm. The objection that
follows is that the tax is bought with compute rather than with
grounding. \textbf{This subsection is a post-hoc audit, not one of the
registered controls}; it was run after the fact, in response to that
objection, and is labelled accordingly.
\texttt{out/compute\_confound.txt}, from
\texttt{code/compute\_confound.py}, over \textbf{444 cells} carrying
per-arm token counts.

{\def\LTcaptype{none} %
\begin{center}
\small
\begin{tabular}{@{}lrrr@{}}
\toprule\noalign{}
quantity & median & p90 & max \\
\midrule\noalign{}

total tokens, full / strict & \textbf{1.17x} & 1.60x & 2.72x \\
completion tokens per call, full / strict & \textbf{1.07x} & 1.60x &
---\\
\bottomrule
\end{tabular}
\end{center}

}

Calls per item are equal in every cell but eight, all of them Qwen3
cells at the 16384-token budget, where a stage occasionally returns
nothing and the average falls from 4.00 to between 3.48 and 3.98. Both
arms of a cell have the same stages by construction; the tax is not a
stage-count difference.

Across cells, the correlation between the token ratio and the tax is
\textbf{r = +0.541}, accounting for roughly \textbf{29\%} of the
variance, and it holds within the headline shape (r = +0.559) and within
MATH-500 (r = +0.723), so it is not an artefact of pooling. A
correlation cannot say which account it supports, because the full arm's
extra tokens are the re-injected problem itself: more tokens and more
information arrive together by construction. On GSM-Hard, a placebo
separates the two. It masks the problem's words to \texttt{\_\_\_},
keeping structure and word count, carries at least \textbf{60\%} of the
extra tokens with at most one word of the problem, and recovers
\textbf{-0.0100 / -0.0050 / -0.0100}. Prompt length and generation
length change with the intervention, and on GSM-Hard length without the
problem's content buys nothing. MATH-500 has no placebo arm.

What it does not rule out: a compute-matched control in which the strict
arm is given the same token budget to spend on something else. That
experiment is not in this paper, and Appendix~H says so.

\subsection{F The sealed held-out test, and the rule it
refuted}\label{b-the-sealed-held-out-test-and-the-rule-it-refuted}

\texttt{PREREG-R16.md} section 6; run files
\texttt{3-runs/r\_*\_HO1\_cross\_*.json} and
\texttt{r\_*\_HO2\_formal\_*.json}.

The rule under test predicted the tax at an interface from two bits:
what the sending edge is instructed to carry, and whether the receiving
stage is permitted to re-derive. It predicted that a \emph{committed}
receiver pays and a \emph{re-deriving} receiver does not. Section 6 of
the registration sealed seven predictions, four of them on two shapes
built to decide it, \texttt{HO1\_cross} and \texttt{HO2\_formal}, and
froze their classification before either had ever run, together with the
rule for reading the outcome: of the four predictions where this rule
and the interface-type-alone rival disagree, losing two refutes it.

\textbf{\texttt{HO1\_cross}} puts two structurally identical condensing
edges in one pipeline, carrying the same \texttt{list\ x\ condense}
message and differing only in the receiver. The rule says stage 1 is
taxed and stage 3 is not.

{\def\LTcaptype{none} %
\begin{center}
\small
\begin{tabular}{@{}lll@{}}
\toprule\noalign{}
model
 & STAGE\_TAX(1), committed
 & STAGE\_TAX(3), re-deriving
 \\
\midrule\noalign{}
Qwen2.5-7B & +0.0800 {[}+0.020, +0.140{]} & \textbf{+0.2700} {[}+0.195,
+0.345{]} \\
gemma-3-12B & +0.0900 {[}+0.030, +0.150{]} & \textbf{+0.4600} {[}+0.390,
+0.530{]} \\
Llama-3.1-8B & +0.0100 {[}-0.050, +0.070{]} & \textbf{+0.2450}
{[}+0.175, +0.315{]} \\
phi-4 & +0.0300 {[}-0.010, +0.070{]} & +0.0350 {[}+0.000, +0.070{]}\\
\bottomrule
\end{tabular}
\end{center}

}

S1 (stage 1 taxed on at least three of four) holds on \textbf{2 of 4}
and fails. S2 (stage 3 null) holds on \textbf{0 of 4} and fails, while
the rival's prediction that stage 3 \emph{is} taxed holds on \textbf{4
of 4}. S3, the ordering the rule exists to produce, is \textbf{reversed
on every model}, with stage 3 taxed at least three times as much as
stage 1 on three of them. \textbf{\texttt{HO2\_formal}} makes every edge
sufficient, so the rule predicts no tax anywhere; the measured
\texttt{full\ -\ strict} is \textbf{-0.0850, +0.0600, +0.0950, +0.1050},
and S4 holds on 0 of 4. S5 predicted no tax on \texttt{bool\_eval}, one
of the synthetic CASCADE tasks of Appendix~H, where the rival predicts
one, and lost as scored, on three cells of which one parses below the
0.50 floor of Appendix~D; S6, that \texttt{logic\_deduction} would
carry the largest tax, was not scored; S7, that the R1-distilled
receiver pays nothing at stage 1, was void. The four decisive
predictions are S2, S4, S5 and S7, and losing two refutes the rule: S2,
S4 and S5 lost, and S7 was void. S2 and S4, both on GSM-Hard, meet the
two-loss rule on their own.

The per-stage picture inside \texttt{HO2\_formal} is sharper than its
total. On Llama-3.1-8B and phi-4 the \texttt{format} stage alone carries
\textbf{+0.1400} (p \textless{} 1e-4 on both) --- and format's incoming
edge is the one the classifier calls immaterial. The rule says
re-grounding it cannot help. Re-grounding it buys fourteen points.

\textbf{What the data supports instead, with its confound named.} In
\texttt{HO1\_cross} the peak stage tax sits at stage 3, \texttt{solve},
on all four models, and it is not simply that later is better: stage 4,
\texttt{format}, sits after stage 3 and is taxed less on every model
(+0.115 against +0.270, +0.180 against +0.460, +0.135 against +0.245,
+0.030 against +0.035). The reading the data supports is that the tax
lives at the stage doing the substantive reasoning rather than at the
stage whose receiver is forbidden to reason. \texttt{HO1\_cross} cannot
separate that reading from a second one, that a re-deriving receiver
pays, because \texttt{solve} is both the reasoning stage and a
re-deriving receiver. The committed setup stage, which also does
substantive work, is taxed +0.0800 on Qwen2.5-7B and +0.0900 on
gemma-3-12B, three to five times less than \texttt{solve}. Amendment 7
registered the test that holds the stage fixed and varies only the
receiver; it is reported below.

\textbf{Two things this does not touch.} The tax itself is measured on
the primary shapes, not on these two, and Section 3.1's family is
unchanged. And the receiver factor of the depth-4 square (Figure 3a,b)
is an effect measured in a square, not a rule for predicting which stage
pays; the square stands and the rule does not. A registered within-model
test of the receiver alone also failed (Amendment 7): on GSM-Hard,
making stage 1 of the four-stage condensing pipeline re-deriving was to
lower its stage tax by at least 0.05 on 3 of 4 models, and it did on 0
of 4; the companion prediction, that the re-deriving stage carries no
tax, held on 2 of 4. They were never the same claim, and conflating them
would understate one and overstate the other.

\textbf{The mixed-order results in full.} In \texttt{MIX\_LR}, where the
loss is between stages 0 and 1 and \emph{both} stages are downstream,
the two help about equally; our registered prediction that stage 1 would
help more failed: it held on 2 of 4 models on each benchmark in the 2024
cohort and on 0 of 3 in the 2026 cohort. The registration's reading of
that failure is that the result is about the stage's own prompt or its
position, not about which interface precedes it; stage 2 is the same
\texttt{solve} prompt in both shapes. Stage 1 is not the \texttt{solve}
stage in either shape; re-grounded, it helps on 7 of 7 models on
MATH-500 after the loss (\texttt{MIX\_LR}) and hurts on 7 of 7 before it
(\texttt{MIX\_RL}). The registration named the \texttt{MIX\_RL} test the
decisive one, and it passes on 7 of 7. The registered prediction that
the total tax does not depend on the order of the two interfaces holds
on GSM-Hard. On MATH-500 it fails for the 2024 cohort, within 0.05 on
only 1 of 4, and its re-test passes for the 2026 cohort at the
registered bar of 2 of 3; there an early loss costs more on 6 of 7
(Figure 4d).

\textbf{On \texttt{HO2\_formal}'s place in the corpus.} Both held-out
shapes sit in the exploratory family, not the primary one (registration
Amendment 24). \texttt{HO2\_formal}'s MATH-500 cells also fail the
registered distinct-response gate, which asks for at least 180 distinct
messages at every stage. Read over the intermediate stages, as in
Appendix~D, its strict arm reaches 124 to 171 and its full arm 122 to
172. On GSM-Hard, where the sealed test ran, three of its four models
pass that gate; Llama-3.1-8B falls to 179 distinct messages at its third
stage, and all four cells are scored above. The shape's numbers are
reported in full so that its placement need not be taken on trust.

\subsection{G What the tax is made of changed on MATH-500; the
prescription did
not}\label{b-what-the-tax-is-made-of-changed-on-math-500-the-prescription-did-not}

\textbf{The drop, and how the registration reads it.} Three registered
tests bear on why the 2026 cohort pays less, and all three pass
(\texttt{out/amend12\_score.txt}). Across the 18 models the test names,
strict-arm accuracy and the tax have a Spearman correlation of
\textbf{$-$0.6006}, against a registered bar of $-$0.60; Qwen3 pays
\textbf{0.1675} less than Qwen2.5; and the 2026 cohort pays
\textbf{0.1067} less than its three predecessors. Amendment 13 had
assigned the two generation tests to re-runs of the truncated cells at a
larger budget, and the scores above take both families at 4096 tokens, a
rule fixed after that amendment. Both tests also pass on the re-runs:
the Qwen3 drop survives as T38 (Amendment 15), and Llama-3.1-8B, the one
model in the two trios that was re-run, pays \textbf{+0.2400} at 8192
tokens, which keeps the cohort drop above 0.10.

The registration fixed two readings in advance. If all three tests
passed, the drop would be written as one mechanism: newer models comply
better under decomposition and so lose less. Strict-arm accuracy is that
reading's only measure; we did not measure compliance on the newer
models. If the depth-4 edge test failed, what shrank between cohorts
would not be the same phenomenon, and the drop would compare two
different things. That test fails, as shown next, so both registered
readings are in force: the drop reads as better compliance under
decomposition, measured here only as strict-arm accuracy, and on
MATH-500 it compares taxes made differently in the two cohorts.

We registered the invariance of the tax's make-up as a prediction, and
it failed from two directions. On MATH-500, the 2026 cohort's edge
effect is positive on only 2 of 3 models, and on gemma-4-12B it is
\textbf{+0.0125} while the receiver effect is \textbf{+0.1425}, the
reverse of its own direct predecessor. Separately, a registered
prediction required the depth-3 square's content-and-form interaction to
reach \textbf{+0.05} on two of the three 2026-cohort models; per model
it is \textbf{+0.0275}, \textbf{+0.0300} and \textbf{-0.0050}, so that
prediction fails too (\texttt{out/amend1516\_score.txt}).
\textbf{Granite-4.1-8B survives Benjamini-Hochberg and not Holm}; the
registered test it bears on, T35, counts Benjamini-Hochberg survivors,
so no registered verdict changes. Figure 5a's cohort means, 0.27 against
0.23, pool a larger 2024 roster and are descriptive. The reading that
survives is narrower than a single-generation study would have licensed:

\begin{quote}
\textbf{The tax persists across the two cohorts. On MATH-500, what it is
made of does not.} For the 2024 cohort it is mostly about the edge; for
gemma-4-12B it is mostly about whether the receiving stage is permitted
to re-derive.
\end{quote}

That distinction is invisible to anyone measuring only the total, which
is what a study on one model generation would have reported, and it is
the reason the prescriptions are stated as interface choices rather than
as a theory of why models fail.

\section{H Limitations}\label{limitations}

\textbf{1. The pipeline loses to a single call, and the monolith
baseline is not clean.} Both are true and the paper reports both. Across
the \textbf{444} distinct three-arm cells,
\texttt{acc(full)\ -\ acc(mono)} averages \textbf{-0.0701} and the
re-grounded pipeline wins \textbf{127} of them; on the headline shape
\texttt{A\_asset} it is \textbf{-0.1201}, winning 17 of 104
(\texttt{out/mono\_gap.txt}). Section 3.4's table prints the arithmetic
for the three current-generation models directly. \textbf{Decomposing
and then repairing it does not beat not decomposing.}

Read from the other end, the same data says how much of that loss sits
at the interfaces. Of the gap between a strict pipeline and a single
call, re-grounding buys back a \textbf{median 63\%}: 63\% over all 322
cells where the monolith leads strict by more than 0.05, 65\% on
\texttt{A\_asset}, 64\% on the two real benchmarks, and 63\% on the
headline cell, where 0.150 becomes 0.555 against a monolith of 0.795.
The fraction is stable across these four slices: \textbf{roughly two
thirds of what a strict pipeline loses to a single call is interface
loss that re-grounding recovers; the remaining third is what a fully
re-grounded pipeline still loses to a single call.}

The budget caveat stands and bounds how far the mono column can be
pushed. Of the fourteen (model, task) pairs run at more than one
completion budget, only \textbf{three} reach a budget where the monolith
truncates at most 10 of 200 items and the strict arm parses at least
0.98; raising the cap trades one failure for the other, and Qwen3-4B on
MATH-500 goes from 59 truncations at 4096 tokens to 15 at 16384 while
\texttt{parse\_rate} falls from 1.000 to \textbf{0.875}
(\texttt{out/budget\_sweep.txt}). So the monolith numbers are a bound
rather than a clean baseline -- which is a reason to read them
carefully, not a reason to withhold them. \textbf{None of it touches the
tax}, which is \texttt{acc(full)\ -\ acc(strict)} between two pipelines
of identical depth, with the monolith in neither arm.

\textbf{2. The tax's make-up on MATH-500 is scoped to the 2024 cohort;
the prescription is not.} Both registered predictions about the tax's
size in the 2026 cohort pass (those models still pay, and pay less), but
on MATH-500 the decomposition of that tax into edge and receiver factors
does not carry over (\S3.4). We report the make-up as a property of the
models we measured it on. Whether the 2026 pattern is itself stable is a
question for a corpus that does not exist yet.

\textbf{3. \texttt{MIX\_LR} and \texttt{MIX\_RL} are not perfectly
matched.} The second stage reads a different kind of input in each
order, so the comparison is not a clean single-factor manipulation and
the ``early loss costs more'' reading is \textbf{descriptive}. What
survives that caveat is the sign result, which does not depend on the
matching: re-grounding the stage upstream of the loss is negative on
\textbf{7 of 7} models on MATH-500 (3 of 7 on GSM-Hard).

\textbf{4. \texttt{HO2\_formal} is not in the primary family, and its
MATH-500 cells fail a registered gate.} On MATH-500 its middle stages
collapse 200 distinct problems into 124 to 171 distinct messages in the
strict arm and 122 to 172 in the full arm, so downstream stages answer a
question that has already lost the item; on GSM-Hard three of four
models pass. We report \texttt{distinct\_per\_stage} for every shape so
that this is checkable rather than asserted.

\textbf{5. Not every effect in the square is resolvable in a single
cell.} At n = 200 one cell resolves about +-0.0550, and FORM (+0.0517)
and the content-by-form interaction (+0.0550) do not clear that line.
They rest on a sign that most of the nine models share (FORM is positive
on 7 of 9) and on pooling, which is weaker evidence than the headline
tax (+0.4050) enjoys. A larger n per cell, not more cells, is what would
settle them.

\textbf{6. There is no compute-matched control.} In Appendix~E the
token ratio across 444 cells accounts for 29\% of the tax's variance, a
correlation that cannot separate tokens from content, and the GSM-Hard
placebo shows that at least 60\% of the added tokens, with at most one
word of the problem, buys nothing. What neither shows is what happens if
the strict arm is handed the full arm's token budget to spend on
something else: more reasoning, a second attempt, a longer answer. That
is the cleanest form of the objection and it is not answered here.

\textbf{7. The synthetic CASCADE tasks (\texttt{logic\_deduction},
\texttt{object\_swap}, \texttt{bool\_eval}) do not measure
decomposition.} We built them, registered them, ran them, and then
retracted them: their apparent effect is instruction/task mismatch, and
\texttt{parse\_rate} together with \texttt{distinct\_per\_stage}
distinguishes the two for free. They stay in the corpus, in the
EXPLORATORY family, as evidence about what the measure does when the
task is wrong -- which is a thing a reader should be able to see.

\textbf{8. Both real benchmarks are mathematical, and the one
non-mathematical task we could measure cleanly says the tax is smaller
there.} GSM-Hard and MATH-500 are both mathematical reasoning, so the
obvious question is whether this is a decomposition tax or a
mathematics-pipeline tax. It cannot be asked with \texttt{A\_asset},
whose intake stage asks for ``the numerical quantities'' and whose setup
stage asks for ``the arithmetic expression(s)'' -- running that on a
non-numeric task measures a mismatch. The domain-agnostic depth-4 shapes
exist for this, and on LastLetter, a verifiable non-mathematical task,
the paired within-model comparison is:

{\def\LTcaptype{none} %
\begin{center}
\small
\begin{tabular}{@{}llrr@{}}
\toprule\noalign{}
model & shape & MATH-500 & LastLetter \\
\midrule\noalign{}

Qwen2.5-7B & \texttt{L4\_lc\_ag} & +0.2700 & \textbf{+0.0750} \\
Qwen3-4B & \texttt{L4\_lc\_ag} & +0.1700 & \textbf{+0.0050} \\
Qwen2.5-7B & \texttt{L4\_pp\_ag} & +0.1550 & \textbf{-0.0500} \\
Qwen3-4B & \texttt{L4\_pp\_ag} & +0.1850 & \textbf{+0.0750}\\
\bottomrule
\end{tabular}
\end{center}

}

Smaller outside mathematics on \textbf{4 of 4} comparisons, over
\textbf{2 models}. That is a direction and not a result, and we report
it as one: \texttt{out/xdomain\_score.txt}.

\textbf{Why only two models, and what the exclusions are.} Six models
ran on LastLetter; the rest of the roster did not. Qwen3-8B can do the
task and has a clean preserving cell (parse \textbf{0.965}, tax
\textbf{+0.1150}, monolith \textbf{0.910}), but its MATH-500 partner
never ran, so it has no pair, and its condensing cell parses
\textbf{0.735}. The other three failed the instrument rather than the
hypothesis, and the monolith arm, which every cell already runs, decides
which. Ministral-8B parses \textbf{0.000} of its pipeline output on
LastLetter, and its \emph{monolith} scores \textbf{0.085}: it cannot do
the task in a single call either, so it cannot inform a decomposition
cost on it. R1-Distill-Llama-8B is the same case at a monolith of
\textbf{0.055}. gpt-oss-20B is the opposite and the more interesting
one: its monolith scores \textbf{0.925}, so the model can do the task,
but its pipeline arm parses \textbf{0.650} -- here the pipeline really
does destroy the answer format, which is a cost, but a different cost
from the accuracy tax and not one this paper measures. MMLU is excluded
outright: the condensing interface discards the answer options along
with everything else, so the final stage cannot name a letter it never
saw, and the strict arm parses \textbf{0.295}. The criterion we apply
throughout is the one those cases share -- \textbf{a model whose
monolith cannot do a task cannot inform the decomposition cost on that
task} -- and it is checkable from data every cell already produces.

\textbf{What we would want next.} The cleanest missing arm is a
within-model matched-budget monolith on a non-thinking model at high
capacity, which would close limitation 1 for at least part of the
roster; and a second current-generation family beyond the three in \S3.4,
which is the difference between ``the decomposition of the tax changed''
and ``gemma-4 is unusual''.

\section{I Related work in full}\label{related-work-in-full}

\textbf{Decomposition as a method, and where each system sits.}
Least-to-most prompting \citep{zhou2022leastt} and Decomposed Prompting
\citep{khot2022decomp} established the regime this paper measures: split
a hard problem into stages, prompt each one separately, pass the result
along.

They do not all pass the same amount along, and our \texttt{strict} arm
is the lossy end rather than the common case. \textbf{Decomposed
Prompting} is the closest: its sub-task handlers receive their sub-query
and not the original question, though a decomposer retains the question
throughout. \textbf{Least-to-most} is further away. It solves
subproblems in sequence and, in its own words, ``solving each subproblem
is facilitated by the answers to previously solved subproblems'', so
every prior answer travels forward. \textbf{PAL} is further still: it is
a single LLM call that emits a program, with the solving step offloaded
to a Python interpreter, so no second LLM stage reads a first stage's
output at all. For \textbf{AutoGen} we could not establish the
visibility model from the paper and make no claim about it.

Two things follow. Our \texttt{strict} arm removes both the original
problem and the earlier stages' outputs, keeping only the immediately
previous one; least-to-most removes neither. So TAX as we measure it
contains the loss from dropping history as well as the loss from
dropping the problem, and this paper does not separate them. What
\texttt{strict} models is the regime a pipeline falls into when a
context budget forces it to carry one message between stages, which we
take to be common in deployed systems and do not demonstrate here. PAL
\citep{gao2022pal} states the premise we test almost exactly, that
models decompose reliably and fail in the solving step, and it is the
paper that contributed GSM-Hard, one of our two benchmarks. Recent work
adds error-recovery machinery to the decomposition itself
\citep{hernndezgutirrez2025recurs}, which presumes a cost worth
recovering from. Our contribution is not a better decomposition method:
it is the finding that the loss is not where the method papers assume,
and an address for it.

\textbf{The closest prior measurement, and what it confounds.}
\citet{radhakrishnan2023questio} compare chain-of-thought prompting,
\emph{CoT decomposition} (subquestions and their answers generated in
one context) and \emph{factored decomposition} (each subquestion
answered in a new context), and report \textbf{86.0}, \textbf{85.6} and
\textbf{81.8} accuracy averaged over four QA tasks on one Claude-family
model (their Table 4). The \textbf{3.8}-point drop between their last
two arms is the closest published quantity to our tax, and it is prior:
that answering in an isolated context costs accuracy was measured, and
named, in 2023. Two things separate it from what we report. Their two
decomposition arms differ in \emph{two} ways at once, because moving
from CoT decomposition to factored decomposition changes both the number
of sampling calls and what each call may see, where ours differ in one
with the call structure frozen. And their design contains no arm that is
decomposed into separate calls \emph{and} re-grounded, which is the arm
that wins here. What this paper adds to their 3.8 points is a
measurement on open-weight models with one factor changed, an address
for it, and a repair with a direction. Their strongest arm is the
monolith at \textbf{86.0}, and so is ours: across our corpus the
re-grounded pipeline still loses to a single call. We agree with them
about the ranking and differ about which quantity is interesting: of the
gap a strict pipeline opens against a single call, re-grounding every
stage recovers a median \textbf{63\%}, and the third that remains is
what a fully re-grounded pipeline still loses to a single call
(Limitation 1).

\textbf{Who has already priced a decomposition, and in which currency.}
Three peer-reviewed results bear on the quantity rather than on the
method. AgentAsk \citep{lin2026agentask} localises multi-agent failure
at the inter-agent message handoff and supplies an edge-level taxonomy
of it --- data gap, signal corruption, referential drift, capability gap
--- repaired by a learned module that asks a clarifying question at the
edge, for up to 4.69\% accuracy at under 10\% added latency. Every
category there is an \emph{error} on the edge; our tax is what the same
edge costs when nothing on it is wrong, since no fault is injected in
any arm, and our repair adds no agent, no call and no question --- only
the problem back into an existing stage's prompt. \emph{Decomposition
Dilemmas} \citep{hu2025decompo} measures the same trade-off we do, in
fact-checking: accuracy gained by decomposing against noise introduced
by it. DnDScore \citep{wanner2025dndsco} then verifies each decomposed
subclaim back in the context it was decomposed out of, which is our
\texttt{full} arm published as a method --- the difference being that
they propose it as a fix and we use it as an instrument, to price what
its absence costs.

\textbf{Prompt-format sensitivity.} \citet{sclar2023quanti} is the
closest prior result to our FORM main effect: meaning-preserving
formatting changes move accuracy by as much as 76 points, which is
larger than anything we report. Two things differ. Their variation is
applied to a whole prompt, where ours is applied to a \textbf{single
interface inside a pipeline} with every other stage held byte-identical;
and their design has no content arm, so format sensitivity and content
loss cannot be separated. Our depth-3 square crosses the two
deliberately, and the answer is that in the 2024 models they compound
(interaction positive on all six 2024-cohort models and \textbf{+0.0550}
over all nine, the same order as either main effect) rather than one
explaining the other. \citet{tang2025large} report format beating
descriptions on a different axis, and \citet{sun2025tables} study
structured intermediates as a benefit; our list-versus-prose result is
the same knob read as a \textbf{cost}, which is what a pipeline exposes
and a single prompt does not.

\textbf{Multi-agent failure taxonomies.} Multi-agent debate
\citep{du2023improv} is the best-known argument that splitting a problem
across calls \emph{buys} something, and it is the premise this paper
prices: debate spends several calls to improve an answer, where a
pipeline spends several calls to produce one, and only the second design
pays at an interface. MAST \citep{cemri2025why} derives 14 failure modes
from 1600+ traces at kappa = 0.88, and MAS-FIRE \citep{jia2026masfir}
injects faults into multi-agent systems to measure reliability. Closest
on the intervention side is \citet{huang2025resili}, who inject errors
into agents' responses and report that adding an Inspector agent to
review and correct messages recovers up to \textbf{96.4\%} of them.
Their Inspector is handed the chat history, so it is grounded in our
sense; what their design cannot do is separate \emph{a reviewer helps}
from \emph{a reviewer holding the original problem helps}, because what
the Inspector sees is never varied. That separation is what our
grounding arms perform, and it is why the same intervention appears here
as an instrument rather than as a defence. These are the observational
counterpart to what we do: they catalogue failures in the wild across
many uncontrolled differences, and we manipulate one edge at a time with
the rest of the pipeline frozen. The two lines agree where they can be
compared. MAS-FIRE and \citet{yadav2026more} both report that
\textbf{stronger foundation models do not uniformly improve robustness}
-- independent arrivals at the shape of our generational result, from
failure injection and from cooperation games respectively, where ours
comes from a registered prediction that we watched fail.

\textbf{Context loss and repair.} \emph{Lost in the Middle}
\citep{liu2024lost} is the standing account of context degradation, and
it is \textbf{positional}: information is present but poorly used
according to where it sits. Our placebo and dose-response arms are what
separate that mechanism from ours. A placebo carrying at least 60\% of
the tokens and at most one word of the problem recovers nothing
(\textbf{-0.0100 / -0.0050 / -0.0100}), and the recovery curve is flat
through half the problem, with 15\% to 40\% recovered at three quarters,
so the effect tracks \emph{what is missing}, not \emph{where anything
sits}. On the repair side, R3 prompting \citep{tian2023r} re-grounds a
degraded context and adaptive injection decoding \citep{jin2025well}
intervenes mid-generation; both are interventions of the kind our
per-stage grounding arms apply, but applied to improve a system rather
than to localise a defect within it. We use grounding as an instrument:
re-grounding exactly one stage is what makes STAGE\_TAX(i) an
attribution rather than an improvement. The closest prescription to ours
comes from a different object: \citet{kwon2026reclaim} compresses a
model's \emph{memory} at a fixed budget and finds that what survives
decides whether a later correction can be acted on -- keep the
recomputable source, drop the conclusion that can be re-derived from it
-- defended, as ours is, with a length-matched control. Ours is a
message between two stages rather than a memory across turns, and our
outcome is clean accuracy rather than recovery from a delivered
correction, but the two prescriptions rhyme: what an interface keeps
matters more than how much of it there is.

\textbf{Test-time compute.} A pipeline spends more inference compute
than a single call. A large literature studies how to spend it well:
compute-optimal allocation \citep{snell2024scalin}, inference scaling
laws across strategies at matched budget \citep{wu2024infere}, the same
allocation posed as an explicit optimisation problem
\citep{wang2025genera}, and allocation across the heterogeneous subtasks
of a multi-stage task \citep{wang2025agentt}. AgentTTS states plainly
that prior test-time scaling work is single-stage, which is also our
starting observation. \textbf{The difference is what is held fixed.}
That literature varies the budget and asks how best to spend it; we hold
the design fixed (same stages, same calls, same budget) and vary one
interface, then ask what it cost. Treating multi-LLM collaboration as an
optimisable graph \citep{wang2025genera} makes the same object we
measure, the edge between two stages, into a search space; our
contribution is the price tag on an edge, which is what a search over
edges would need and does not have.

\textbf{Error propagation, arrived at from elsewhere.} The closest
external result to ours comes from software engineering, not from
reasoning research. \citet{konstantinou2026on} study LLM-generated test
suites and find that a stage which reads a previous stage's output does
markedly worse than one that works independently: fault detection falls
from \textbf{25\% to 14\%} when tests are generated after the code
rather than beside it. Different field, different task, no contact with
this work, and the same structure as the tax. We take that as support
rather than competition, and it also sharpens what is ours: they
encountered the effect as a confound to be avoided in a study about
something else, and so did not ask what the loss is \emph{about}, which
is the question Section 3.2 and Section 3.3 answer.

On the attribution side, DisasterBench \citep{chen2026disast} introduces
First-Point-of-Failure to localise the earliest root cause in a tool-use
pipeline and separate primary errors from downstream cascading ones.
That is the same problem \texttt{STAGE\_TAX(i)} addresses, approached
from the opposite direction: they \emph{infer} the origin from an
observed trajectory, while we \emph{intervene} on one stage at a time
and read the recovery. Inference is cheaper and applies to traces that
already exist; intervention is what licenses a causal claim about which
stage the tax belongs to. CORRECT \citep{yu2025correc} builds the same
kind of evidence at scale from 2,000+ injected-failure trajectories.

\textbf{Statistics.} We report McNemar's paired test
\citep{mcnemar1947note} per cell, and correct within declared families
using Holm's step-down procedure \citep{holm1979a} and
Benjamini-Hochberg \citep{benjamini1995contro}. \citet{dror2020statis}
is the NLP-specific standard our family split is measured against, and
its central warning, that a corpus-wide correction over heterogeneous
hypotheses answers a question nobody asked, is why \S3.1 corrects within
families at all. Membership is fixed by each cell's manifest and never
by its outcome.

\end{document}